\documentclass[runningheads]{llncs}
\usepackage[T1]{fontenc}
\usepackage{float}
\usepackage{pifont}
\usepackage{footnote}
\usepackage{enumitem}
\usepackage{bm}
\usepackage{arydshln}
\usepackage{booktabs}
\usepackage{multicol}
\usepackage{multirow}
\usepackage{color}
\usepackage{xcolor}     
\usepackage{colortbl}
\usepackage{soul}
\usepackage{bbding}
\usepackage{makecell}
\usepackage{mathtools}
\usepackage{imakeidx}
\usepackage{amssymb}
\usepackage{graphicx}
\usepackage{amsmath}
\usepackage{threeparttable}
\definecolor{citecolor}{HTML}{0071BC}
\definecolor{linkcolor}{HTML}{ED1C24}
\usepackage[colorlinks,
            anchorcolor=red,
            citecolor=citecolor, 
            linkcolor=linkcolor,
            ]{hyperref}
\makeindex
\usepackage{arydshln}
\usepackage{lipsum}
\usepackage[toc]{multitoc}
\usepackage[edges]{forest}
\usepackage[normalem]{ulem}

\usepackage{bbding}
\usepackage[most]{tcolorbox}

\usepackage{algorithm}
\usepackage{algorithmic}

\usepackage{minitoc}
\usepackage[toc,page,header]{appendix}

\definecolor{orchid}{rgb}{0.85, 0.44, 0.84}
\definecolor{rubinred}{rgb}{0.82, 0.0, 0.28}
\definecolor{flagship}{rgb}{0, 255, 0}
\definecolor{radiologist}{rgb}{0.50, 0.50, 1}

\definecolor{YT}{HTML}{002FA7}

\newcommand{\method}{RT-Super}

\newcommand{\numofct}{34,152}
\newcommand{\numofmasks}{767}

\newcolumntype{P}[1]{>{\centering\arraybackslash}p{#1}}
\newlength\savewidth
  
\usepackage{graphicx,verbatim}
\usepackage{array}
\newcolumntype{C}[1]{>{\centering\arraybackslash}m{#1}}

\usepackage[dvipsnames]{xcolor}
\definecolor{lightorange}{RGB}{255,230,204} % define your own light-orange
\begin{document}
%
% \title{Learning Longitudinal Multi-Tumor Segmentation from Radiology Reports}
% \title{Learning Multi-Cancer Segmentation from Reports and Longitudinal Data}
\title{RT-Super: Learning Tumor Segmentation from Longitudinal Images and Reports}
\titlerunning{\method}
% \begin{comment}
%index{Bassi, Pedro R. A. S.}
%index{Li, Wenxuan}
%index{Gu, Hanxue}
%index{Chen, Jieneng}
%index{Zhou, Xinze}
%index{Zhu, Zheren}
%index{Er, Sezgin}
%index{Hamamci, Ibrahim E.}
%index{Menze, Bjoern H.}
%index{Akan, Gulhan E.}
%index{Wang, Kang}
%index{Yang, Yang}
%index{Yuille, Alan L.}
%index{Zhou, Zongwei}
\author{
Pedro R. A. S. Bassi\inst{1,2,3} \and
Wenxuan Li\inst{1,2,3} \and
Hanxue Gu\inst{4} \and
Jieneng Chen\inst{1} \and
\\Xinze Zhou\inst{1} \and
Zheren Zhu\inst{4} \and
Sezgin Er\inst{5} \and
Ibrahim E. Hamamci\inst{5} \and
\\Bjoern H. Menze\inst{5} \and
Gulhan E. Akan\inst{6} \and
Kang Wang\inst{4} \and
Yang Yang\inst{4} \and
\\Alan L. Yuille\inst{1} \and
Zongwei Zhou\inst{1,7}\thanks{Correspondence to: Zongwei Zhou (\href{mailto:zzhou82@jh.edu}{\texttt{zzhou82@jh.edu}})}
}
\authorrunning{P. Bassi et al.}
\institute{
Johns Hopkins University \and
Harvard Medical School \and
Massachusetts General Hospital \and
University of California, San Francisco \and
University of Zurich \and
Istanbul Medipol University \and
Johns Hopkins Medicine
}
% \end{comment}

% \author{Anonymized Authors}  %% Added for anonymized MICCAI submission
% \authorrunning{Anonymized Author et al.}
% \institute{Anonymized Affiliations}

\maketitle              % typeset the header of the contribution
\begin{abstract}

Multi-tumor segmentation is important for early cancer detection and allows radiologists to visualize, verify, and understand AI predictions. However, tumor segmentation masks are expensive, time-consuming, and unavailable for many tumor types in public data. Instead, hospitals have vast, readily available data that can guide segmentation: radiology reports, longitudinal images, and multi-phase images. We use this readily available data to substitute for tumor masks in training AI for tumor segmentation. To this end, we propose a new architecture, RT-Super. It has a teacher network, which analyzes the patient's longitudinal images and reports to create high-quality tumor masks. These masks train a student network, which sees a single image and no report. At inference, when longitudinal images and reports are unavailable, we use the student. RT-Super uses a new CNN-Transformer architecture and novel Consistency Losses that exploit tumor location consistency across longitudinal images. We train RT-Super to segment esophagus, uterus and spleen tumors, which have few or no public masks. Even without training masks, RT-Super can segment these tumors and surpass public AI models. Overall, we demonstrate that learning from longitudinal images, multi-phase images, and reports can overcome mask scarcity and advance multi-cancer detection and segmentation. Code: \url{https://github.com/MrGiovanni/RT-Super}.

%Limitation of single image analysis. Dynamics refer to detection, segmentation, and diagnosis without any treatment response. We make three contributions: 1. novel methodology, 2. risk estimation, 3. malignancy identification.

\keywords{Tumor Segmentation \and Longitudinal Data \and Reports}
\end{abstract}
\section{Introduction}\label{sec:introduction}

\setcounter{footnote}{0}

Early detection of multiple cancer types can greatly improve patient survival \cite{bassi2025scaling,cao2023large,li2026early}, but multi-tumor segmentation is challenging: tumor segmentation masks are expensive and time-consuming to create~\cite{chou2024acquiring}, mostly unavailable in hospitals, and public datasets offer masks for only a few tumor types. For example, public computed tomography (CT) datasets and segmentation models mainly cover tumors in the lungs, liver, pancreas, kidneys, and colon~\cite{heller2019kits19,li2025pants,liu2023clip}. 

While tumor masks are scarce, hospitals routinely collect rich alternative data that can guide tumor segmentation: (1) \textit{Longitudinal images}: past and future scans of the same patient, where tumors that were small and subtle in past scans may become larger and clearer in future scans; (2) \textit{Multi-phase images}: CT scans with different contrast phases, where intravenous contrast can improve tumor visibility; and (3) \textit{Radiology reports}: expert-written text often describing tumor size, location, and count. These data sources are readily available in hospitals and can effectively substitute for or augment scarce tumor masks. Thus, we ask: \textit{can AI architectures and training methods exploit reports, longitudinal images, and multi-phase images to learn tumor segmentation with fewer or no tumor masks?}

We hypothesize that leveraging longitudinal images, multi-phase images, and radiology reports in training enables AI to segment multiple tumor types without abundant segmentation masks and improves inference performance with only a single image and no report. This hypothesis stems from our observation that, when asked to segment tumors, radiologists often request all available patient images across time, multi-phase images, and all their reports~\cite{li2026radthinking}. By looking at future scans or contrast-enhanced scans, radiologists can better find and segment tumors that are small or barely visible in previous or non-contrast scans. 

Inspired by this, we propose the \method\ architecture (Report \& Time Supervision). During training, a teacher network receives as input longitudinal and multi-phase images and all their reports. By exploring this comprehensive information, the teacher can create higher-quality tumor masks. In the absence of ground-truth tumor masks, the masks made by the teacher are used to train a student network, which only receives one image and no report. Therefore, the student can be used at inference when longitudinal/multi-phase images and reports are unavailable\footnote{A first diagnostic image often has no prior images; reports are always assumed missing, as AI does not need to detect tumors already detected and described in reports.}. While the student learns from the teacher masks, the teacher learns from Report Supervision losses~\cite{bassi2025learning} and a new Consistency Loss. The new loss exploits consistencies across longitudinal images, so that images where tumors are larger help the teacher segment images where tumors are smaller. For efficiency, the student is part of the teacher network (self-distillation). With \method, we demonstrate that learning from reports, longitudinal images, and multi-phase images improves tumor segmentation even when only a single image and no report are available at inference. 

To train \method, we used a large-scale dataset, with \numofct\ CT scans. This dataset focuses on esophagus, spleen and uterus tumors, because these tumor types have scarce public tumor masks (fewer than 60 for uterus and esophagus~\cite{chen2025scaling}, none for spleen). Our dataset exemplifies what is readily available in hospitals: reports, longitudinal images, and multi-phase images, but no masks. Here, we train both with many reports and no tumor mask, and with many reports and few tumor masks (\numofmasks, created by our collaborating radiologists). In tumor detection, \method\ substantially surpassed public AI models such as Google's MedGemma~\cite{sellergren2025medgemma}, Stanford's Merlin~\cite{blankemeier2024merlin}, and Universal Lesion Segmentation (ULS)~\cite{de2025uls23} (Tab.~\ref{tab:rt_super_results_int}). It also surpassed alternative tumor segmentation methods trained on our dataset, such as CLIP~\cite{blankemeier2024merlin,radford2021learning}, multi-task learning (MTL)~\cite{zhang20213d}, and self-supervised learning (Models Genesis)~\cite{zhou2021models} (Tab.~\ref{tab:rt_super_results_int}). Our main contributions are:
\begin{enumerate}
    \item \method, an architecture that learns multi-tumor segmentation with fewer or no masks. It learns from radiology reports, longitudinal images, and multi-phase images. It improves performance even with a single image and no report at inference (Tab.~\ref{tab:rt_super_results_int}, \ref{tab:rt_super_results_small_combined}).
    \item We created the Consistency Loss, which exploits the consistency across longitudinal and multi-phase images to improve tumor segmentation.
    \item We release \method\ that largely surpassed public AI models in the detection and segmentation of esophagus, spleen, and uterus tumors (Tab.~\ref{tab:rt_super_results_int}, \ref{tab:rt_super_results_small_combined}).
\end{enumerate}

\textbf{Related work.} Previous work explored synthetic tumors~\cite{chen2025scaling,hu2023label} and radiology reports, readily available in hospitals, for learning tumor segmentation. Multi-task learning strategies extract classification labels (e.g., tumor presence or absence) from radiology reports. These labels train the AI model for classification, while the available tumor masks train it for segmentation~\cite{zhang20213d}. Vision-language foundation models (VLMs) were trained on reports with contrastive losses (e.g., CLIP~\cite{radford2021learning}), and fine-tuned for segmentation with masks~\cite{blankemeier2024merlin}. These strategies used reports to learn auxiliary tasks (classification or CLIP). Instead, Report Supervision~\cite{bassi2025learning} used reports to directly supervise segmentation, introducing loss functions that make segmented tumors match the tumor count, sizes, and locations described in reports. This approach substantially improved performance. However, prior work has not used longitudinal and multi-phase CT scans as an additional source of information to further substitute for masks in learning tumor segmentation.

\begin{figure}[t]
    \centering
    \includegraphics[width=1\linewidth]{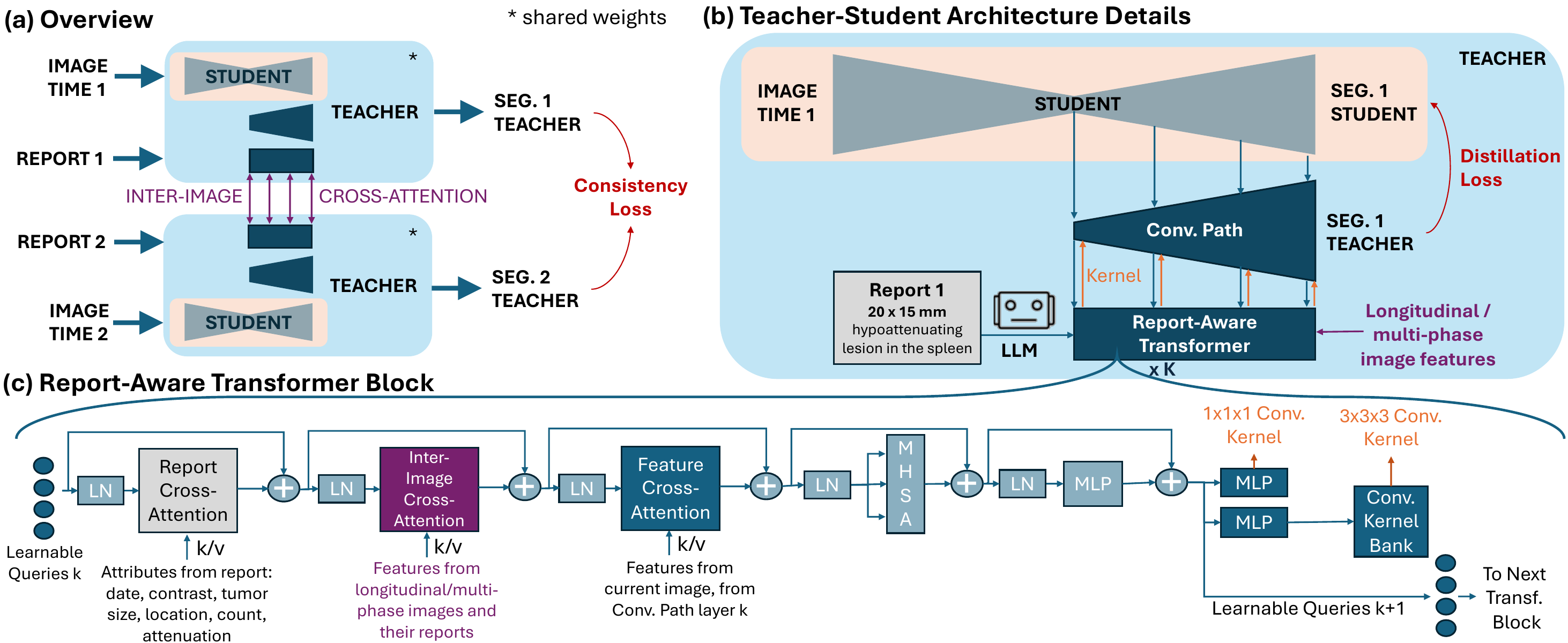}
    \caption{\textbf{Overview of \method.}
    \textbf{(a) Multi-image training.} \method\ is a teacher-student framework. For a patient with multiple scans (e.g., different time points or contrast phases), we run one teacher-student instance (shared weights) per scan. Each student processes one scan independently. The teachers exchange information via inter-image cross-attention and are trained with an additional Consistency Loss across scans.
    \textbf{(b) Teacher-student distillation.} The teacher leverages information from multiple images and from reports to create high-quality tumor segmentation masks. The student sees one image only, but it is trained with the high-quality masks created by the teacher. At inference, when multiple images and reports are often \textit{not} available, we use only the student.
    \textbf{(c) Report-aware transformer.} The teacher has a transformer that receives information from longitudinal images and reports via cross-attention. It uses this information to create convolutional kernels, which refine features from the student to create an improved, report- and longitudinal-aware tumor mask. As the teacher creates this mask by refining the student's features, the student is \textit{part} of the teacher.}
    \label{fig:method}
\end{figure}

\section{\method}\label{sec:method}

\method\ is based on a new self-distillation architecture. The student is a standard segmentation model, of any architecture (here, MedFormer~\cite{gao2022data}). It receives a single image (no report) and segments tumors. The teacher is a hybrid between transformers and convolutional networks. It receives information from radiology reports, longitudinal images, and multi-phase images. The teacher exploits this information to refine the deep features of the student, creating an improved tumor segmentation mask. This mask is used to train the student in the absence of ground-truth tumor masks. The \method\ architecture is shown in Fig.~\ref{fig:method}, and explained in Sec.~\ref{sec:architecture}. To better exploit consistencies across multi-phase and longitudinal images, we introduce Consistency Losses. Before training \method, we use a large language model (LLM) to extract relevant tumor information from reports, including tumor count, locations (organs), and diameters~\cite{bassi2025learning}. The LLM runs only once. We use the Llama 3.1 70B AWQ LLM, which was shown to extract tumor information from reports with 96\% accuracy~\cite{bassi2025radgpt}.

\subsection{\method\ Architecture}
\label{sec:architecture}

\method\ uses a self-distillation, teacher-student architecture, where the student is the initial part of the teacher network (Fig.~\ref{fig:method}). Any segmentation architecture can be used as the student (we used MedFormer~\cite{gao2022data}). The student receives one image and creates its tumor mask. The teacher improves this mask by leveraging information from reports and from longitudinal images. The teacher has a \textit{convolutional path} and a \textit{report-aware transformer}. The convolutional path refines the deep features of the student with a sequence of convolutional layers. The kernels of these layers are created by the report-aware transformer, which has access to the student's deep features, and to information from radiology reports and multiple images. Thus, the transformer creates convolutional kernels that iteratively refine the student's features, finally creating a tumor mask that better matches reports and is more consistent with longitudinal and multi-phase images. This mask is the teacher's output.

\textbf{The teacher's convolutional path} is a sequence of convolutional blocks, one for each resolution level in the student decoder. The teacher's convolutional block k refines the student decoder features at level k, $S_k$. Specifically, the input of the teacher convolutional block k is $S_k$, concatenated with $T_{k-1}$, the output of the teacher's previous convolutional block (up-sampled). Each convolutional block applies a $3\times3\times3$ convolution followed by a $1\times1\times1$ convolution (each with instance normalization and leaky ReLU). Crucially, the kernels of these two convolutions are produced by the teacher's report-aware transformer. After the teacher's final convolutional block, a $1\times1\times1$ convolution (transformer-generated) maps the final refined features to a tumor segmentation map (teacher's output).

\textbf{The teacher's report-aware transformer} (Fig.~\ref{fig:method}) starts from learnable queries. These queries are updated by transformer blocks, encoding information from reports and longitudinal/multi-phase images (explained later). Then, the updated queries are used to create convolutional kernels. In transformer block k, a \textit{report cross-attention layer} updates the queries with cross-attention to tumor attributes extracted from reports (keys/values)\footnote{Attributes: tumor diameter, organ, slice (tumor z coordinate), attenuation (hyper-/hypo-/iso-attenuating), and malignancy. Attributes are encoded numerically.}. Afterwards, a \textit{feature cross-attention layer} updates the queries with cross-attention to the input of convolutional block k (which includes deep features from the student), followed by standard multi-head self-attention and a multi-layer perceptron (MLP). Updated queries at the output of transformer block k are used to create kernels for the convolutional block k. For the $1\times1\times1$ convolution, an MLP directly creates the convolutional kernel from multiple queries%\footnote{We select $C_{out}$ output queries and pass each through an MLP, which maps it to a vector of size $C_{in}$. Stacking the vectors forms the kernel of shape $[C_{out},\, C_{in},\, 1,\, 1,\, 1]$.}
. For the $3\times3\times3$ convolution, directly generating its kernel would be expensive. 
So, we use an approach inspired by soft mixture of experts and Dynamic Convolutions~\cite{chen2020dynamic}: the transformer chooses which $3\times3\times3$ kernel to use, from a learnable bank of $M$ candidate kernels (we use $M=16$). An MLP with softmax activation takes one query from the output of the transformer block k, and predicts mixture weights for this bank. The final $3\times3\times3$ kernel is then computed as the weighted linear combination of the $M$ kernels in the bank.

The transformer blocks are connected sequentially. Each block k updates its input queries. Some of its output queries are used to create the convolutional kernels for the convolutional block $k$; the others are given to the next transformer block, $k+1$. A special transformer block is placed before all kernel-generating blocks. It updates all learnable queries and forwards them to the subsequent blocks. In this update, the block performs cross-attention between the queries and three concatenated segmentation masks: the student's output, a mask for the organ with tumors, and a mask for the tumor slices. The tumor slices and the organs with tumors are defined by the report, and the corresponding organ masks are created before training\footnote{We used nnU-Net~\cite{isensee2021nnu} trained for organ segmentation on AbdomenAtlas~\cite{li2024abdomenatlas}. Public AI models~\cite{bassi2024touchstone,wasserthal2023totalsegmentator} can create masks for multiple organs, but not their tumors. We use 2\,cm of binary dilation to compensate for mask errors.}. These masks spatially inform the report-aware transformer where the student thinks the tumor is, and where the tumor should actually be, according to the report.

\textbf{When longitudinal or multi-phase images are available} for a patient, we run one teacher-student instance per image (shared weights). Each student instance processes only its own image. In contrast, the teacher instances communicate through cross-attention. Specifically, in each teacher transformer block k, we add an \emph{inter-image cross-attention} layer after the report cross-attention (Fig.~\ref{fig:method}). This layer updates the current instance's learnable queries by attending to the \textit{other} teacher instances---to the input features of the convolutional block k in each other teacher instance. %\footnote{For efficiency, we train with 2 images per patient. For patients with more images, we randomly select 2 per training iteration, so that all images are used during training.}. 

We also inform the teacher about the date for each image and its contrast phase, with new tokens in the report cross-attention. Thus, when generating convolutional kernels, each teacher instance can leverage reports together with information from longitudinal and multi-phase images. \method\ handles missing data. When report attributes or longitudinal images are missing, we set them to zero in the report cross-attention. As \method\ targets both tumor segmentation and detection, we add a classifier on top of the teacher and student segmentation outputs. It is trained jointly with the segmenter.

\subsection{\method\ Loss Functions}

For images that have ground-truth tumor masks, we directly use these masks to train the teacher and the student, using the usual Dice and binary cross-entropy (BCE) losses. For images without masks, we train the teacher with Report Supervision losses~\cite{bassi2025scaling}. The Volume Loss teaches the teacher to segment tumors matching the tumor volume and locations (organs) estimated from reports. The Ball Loss makes the segmented tumors match reports in terms of tumor diameter, count, and locations. 

For images without masks, we train the student with distillation: Dice and BCE losses encourage the student's output to match the teacher's output. We do distillation with hard targets, created by binarizing the teacher's tumor segmentation output with report-based post-processing (Ball Loss pseudo-mask~\cite{bassi2025learning}).\footnote{This post-processing refines a soft mask into a binary mask that better matches the tumor size, count, and location in reports.} Through distillation, a student that sees a single image and no report can learn from high-quality masks created by a teacher that exploits privileged information---longitudinal images, multi-phase images, and their reports.

We also propose a \textbf{Consistency Loss} that lets the teacher use longitudinal images with larger, clearer tumors to guide segmentation in scans where tumors are smaller and unclear. As tumors can grow over time or shrink with treatment, we enforce \emph{location consistency}: tumors segmented in a ``small-tumor'' scan S should lie inside the corresponding tumors in a ``large-tumor'' scan L (plus a conservative margin). We apply the Consistency Loss when the radiology report indicates that all tumors in scan S are smaller than all tumors in scan L. 
\textbf{(1)~Registration.} We register scan L to S using uniGradICON~\cite{tian2024unigradicon}, which predicts a deformation field based on images and organ masks\footnote{We fine-tune uniGradICON during segmentation training. If registration fails (Dice similarity coefficient, DSC, $<0.8$ between registered and target organ masks), we skip the Consistency Loss.}. With the deformation field, we register the teacher's output tumor mask, made for scan L, to the space of S. The teacher's output tumor mask is binarized with report-based post-processing before registration, improving the agreement between the tumor mask and the report. We dilate the registered tumor mask by 2\,cm to compensate for registration errors. \textbf{(2)~Consistency.} We enforce that tumors segmented in scan S lie inside the registered and dilated tumor mask of scan L by modifying the Volume Loss and Ball Loss~\cite{bassi2025learning}. Originally, these losses encourage segmented tumors to lie inside the organ where the report mentions tumors, localized with a precomputed organ mask. Here, we replace that organ mask with its intersection with the registered (and dilated) tumor mask from L, and then apply the Ball and Volume losses on S.

\section{Results}

\begin{table}[!t]
\centering
\tiny
\setlength{\tabcolsep}{2pt}
\caption{\textbf{Trained with longitudinal and multi-phase images and reports, \method\ improves multi-tumor detection with a single image and no report at inference (internal test).} \method\ surpasses previous public AI models and alternative training methods. The main gain over R-Super is when training without masks. This dataset includes malignant tumors and control (no-tumor) cases. Sensitivity (Se), specificity (Sp), and F1 at the operating point maximizing balanced accuracy for each model. $^\dagger$: trained with reports only (no tumor masks). Bold: best within each group.}

\begin{tabular}{p{0.185\textwidth}*{3}{>{\centering\arraybackslash}p{0.060\textwidth}}*{12}{>{\centering\arraybackslash}p{0.035\textwidth}}}
\toprule
& \multicolumn{3}{c}{\scriptsize train} & \multicolumn{3}{c}{\scriptsize spleen} & \multicolumn{3}{c}{\scriptsize esophagus} & \multicolumn{3}{c}{\scriptsize uterus} & \multicolumn{3}{c}{\scriptsize average} \\
\cmidrule(lr){2-4}\cmidrule(lr){5-7}\cmidrule(lr){8-10}\cmidrule(lr){11-13}\cmidrule(lr){14-16}
\scriptsize model & longi. & report & mask & Se & Sp & F1 & Se & Sp & F1 & Se & Sp & F1 & Se & Sp & F1 \\
\midrule
\multicolumn{16}{l}{\textit{public AI models}} \\
Merlin~\cite{blankemeier2024merlin} &  & x &  & 0 & 100 & 0 & 0 & 100 & 0 & 0 & 100 & 0 & 0 & 100 & 0 \\
ULS~\cite{de2025uls23} &  &  & x & 28 & 89 & \textbf{34} & 5 & 98 & \textbf{10} & 32 & 85 & \textbf{42} & 22 & 91 & \textbf{29} \\
MedGemma~\cite{sellergren2025medgemma} &  & x &  & 6 & 95 & 10 & 0 & 100 & 0 & 0 & 100 & 0 & 2 & 98 & 3 \\
\midrule
\multicolumn{16}{l}{\textit{trained on our dataset --- report only}} \\
classification$^\dagger$ &  & x &  & 24 & 78 & 24 & 78 & 85 & 84 & 75 & 82 & 75 & 59 & 82 & 61 \\
R-Super$^\dagger$~\cite{bassi2025learning} &  & x &  & 88 & 87 & 72 & 71 & 88 & 80 & 92 & 82 & 82 & 84 & 86 & 78 \\
\textbf{\method$^\dagger$} & x & x &  & 71 & 93 & \textbf{73} & 89 & 80 & \textbf{90} & 87 & 88 & \textbf{84} & 82 & 87 & \textbf{82} \\
\midrule
\multicolumn{16}{l}{\textit{trained on our dataset --- report and/or mask}} \\
CLIP~\cite{blankemeier2024merlin,radford2021learning} &  & x & x & 77 & 81 & 63 & 82 & 85 & 87 & 87 & 80 & 81 & 82 & 82 & 77 \\
MTL~\cite{chen2019lesion} &  & x & x & 77 & 83 & 66 & 76 & 94 & 85 & 92 & 83 & \textbf{85} & 82 & 87 & 79 \\
Models Genesis~\cite{zhou2021models} &  &  & x & 79 & 74 & 58 & 83 & 85 & 87 & 86 & 78 & 79 & 83 & 79 & 75 \\
segmentation~\cite{gao2022data} &  &  & x & 91 & 69 & 62 & 82 & 90 & 88 & 88 & 80 & 82 & 87 & 80 & 77 \\
nnU-Net~\cite{isensee2021nnu} &  &  & x & 68 & 68 & 50 & 79 & 88 & 86 & 80 & 88 & 82 & 76 & 81 & 73 \\
R-Super~\cite{bassi2025learning} &  & x & x & 87 & 90 & 78 & 86 & 94 & \textbf{91} & 86 & 85 & 83 & 86 & 90 & 84 \\
\textbf{\method} & x & x & x & 83 & 95 & \textbf{83} & 88 & 89 & \textbf{91} & 84 & 82 & 80 & 85 & 89 & \textbf{85} \\
\bottomrule
\end{tabular}
\label{tab:rt_super_results_int}
\end{table}

\begin{table}[!t]
\centering
\setlength{\tabcolsep}{1pt}
\caption{\textbf{External validation on small tumors and ablations.} We test \method\ on data from an unseen hospital. Trained with reports only, \method\ substantially surpasses R-Super in segmentation DSC. Detection (Se, Sp, and area under the ROC curve, AUC) is evaluated on small tumors ($<$2\,cm) only; DSC on masks (35\% small). DSC is lower for these tumor types than for others such as pancreatic tumors~\cite{li2025pants}, because these tumor types are harder to segment on CT, which is not their primary imaging modality. Reflecting this difficulty, the DSC between masks created by two radiologists (2 and 8 years of experience) in our test set was 66, 50, and 62 for spleen, esophagus, and uterus tumors, respectively. Sensitivity and specificity at the operating point maximizing balanced accuracy for each model. $^\dagger$: trained with reports only (no tumor masks). Bold: best within each group (ablation rows excluded).}

{\tiny
\begin{tabular}{@{}l*{3}{c}*{16}{c}@{}}
\toprule
& \multicolumn{3}{c}{train} & \multicolumn{4}{c}{spleen} & \multicolumn{4}{c}{esophagus} & \multicolumn{4}{c}{uterus} & \multicolumn{4}{c}{average} \\
\cmidrule(lr){2-4}\cmidrule(lr){5-8}\cmidrule(lr){9-12}\cmidrule(lr){13-16}\cmidrule(lr){17-20}
model & longi. & report & mask & Se & Sp & AUC & DSC & Se & Sp & AUC & DSC & Se & Sp & AUC & DSC & Se & Sp & AUC & DSC \\
\midrule
\multicolumn{20}{l}{\textit{public AI models}} \\
Merlin~\cite{blankemeier2024merlin} &  & x &  & 1 & 99 & 50 & - & 0 & 100 & 50 & - & 33 & 87 & 60 & - & 11 & 95 & 53 & - \\
ULS~\cite{de2025uls23} &  &  & x & 58 & 51 & \textbf{53} & \textbf{4} & 59 & 87 & \textbf{74} & \textbf{11} & 50 & 74 & \textbf{66} & \textbf{12} & 56 & 71 & \textbf{64} & \textbf{9} \\
MedGemma~\cite{sellergren2025medgemma} &  & x &  & 0 & 100 & 50 & - & 0 & 99 & 50 & - & 0 & 100 & 50 & - & 0 & 100 & 50 & - \\
\midrule
\multicolumn{20}{l}{\textit{trained on our dataset --- report only}} \\
classification$^\dagger$ &  & x &  & 22 & 85 & 54 & - & 73 & 72 & 72 & - & 75 & 56 & 72 & - & 57 & 71 & 66 & - \\
R-Super$^\dagger$~\cite{bassi2025learning} &  & x &  & 80 & 85 & \textbf{85} & 27 & 68 & 90 & 82 & 5 & 58 & 83 & 78 & 15 & 69 & 86 & \textbf{82} & 16 \\
\textbf{\method$^\dagger$} & x & x &  & 67 & 91 & 81 & \textbf{31} & 83 & 84 & \textbf{86} & \textbf{33} & 30 & 90 & \textbf{79} & \textbf{25} & 60 & 88 & \textbf{82} & \textbf{30} \\
\midrule
\multicolumn{20}{l}{\textit{trained on our dataset --- report and/or mask}} \\
nnU-Net~\cite{isensee2021nnu} &  &  & x & 60 & 80 & 71 & 26 & 62 & 90 & 78 & 17 & 25 & 89 & 58 & 43 & 49 & 86 & 69 & 29 \\
R-Super~\cite{bassi2025learning} &  & x & x & 73 & 82 & 81 & \textbf{54} & 89 & 80 & 90 & 18 & 83 & 69 & \textbf{80} & 51 & 82 & 77 & \textbf{84} & \textbf{41} \\
\textbf{\method} & x & x & x & 66 & 89 & \textbf{82} & 43 & 89 & 82 & \textbf{91} & \textbf{24} & 75 & 70 & 78 & \textbf{55} & 77 & 80 & \textbf{84} & \textbf{41} \\
\midrule
\multicolumn{20}{l}{\textit{ablation of \method\ --- report and mask}} \\
no Consistency Loss & x & x & x & 62 & 93 & 82 & 44 & 86 & 77 & 89 & 21 & 58 & 78 & 78 & 52 & 69 & 83 & 83 & 39 \\
no inter-image cross-att & x & x & x & 71 & 83 & 81 & 21 & 82 & 86 & 91 & 16 & 70 & 74 & 80 & 53 & 74 & 81 & 84 & 30 \\
no dynamic kernel & x & x & x & 73 & 91 & 83 & 47 & 84 & 87 & 89 & 24 & 75 & 72 & 79 & 48 & 77 & 83 & 84 & 40 \\
no size-info teacher & x & x & x & 69 & 86 & 80 & 45 & 71 & 87 & 86 & 19 & 100 & 59 & 80 & 42 & 80 & 77 & 82 & 35 \\
no report-info teacher & x & x & x & 65 & 87 & 83 & 37 & 86 & 85 & 90 & 18 & 83 & 70 & 77 & 52 & 78 & 81 & 83 & 36 \\
\bottomrule
\end{tabular}}
\label{tab:rt_super_results_small_combined}
\end{table}

\textbf{Datasets. (1) Training:} \numofct\ CT-Report pairs, from 10,385 control patients (10.4\% with longitudinal CTs, 3.8\% multi-phase), 7,597 spleen tumor patients (33.2\% longitudinal, 11.9\% multi-phase), 4,258 uterus tumor patients (23.7\% longitudinal, 8.8\% multi-phase), and 741 esophagus tumor patients (28.7\% longitudinal, 21.3\% multi-phase). The training set includes \numofmasks\ tumor masks (238 spleen, 288 uterus, 241 esophagus). All data were collected from the UCSF hospital and affiliated institutions across California. Longitudinal and multi-phase CT scans are part of the full dataset, and all models were trained with the full dataset. \textbf{(2) Internal Test (pathology-proven):} CT-Report pairs from patients with pathology-proven malignant tumors from UCSF, plus random controls---148 no-tumor CTs, 34 spleen tumor CTs, 235 esophagus tumor CTs, and 84 uterus tumor CTs. \textbf{(3) External Test (small tumors):} CT-Report pairs from an unseen hospital (Istanbul Medipol University, Turkey). Positive scans: spleen 135, esophagus 45, uterus 12; 226 controls. It includes benign and malignant tumors, all small ($\leq$2\,cm in diameter).  \textbf{(4) External Test (masks):} used for DSC calculation, CT-Mask pairs from the same external hospital, spleen 25, esophagus 17, uterus 13. It includes benign and malignant tumors, 35\% of them small. \uline{In testing, AI models have no access to patient reports or to longitudinal images.}

\textbf{Training \method.} We trained two variants: \method$^\dagger$ using CT-Report pairs, and \method\ using CT-Report and CT-Mask pairs. We used MedFormer \cite{gao2022data} as the student, and used its training hyperparameters for \method\ and all MedFormer-based baselines---segmentation, CLIP, Models Genesis, MTL, R-Super, and classification (MedFormer's encoder). We used the R-Super cropping strategy~\cite{bassi2025learning}. Loss weights were 1 for supervised segmentation losses and distillation, and 0.1 for Consistency and Report Supervision losses. By the end of training, the \method\ teacher surpassed its student by 2.9\% DSC. nnU-Net~\cite{isensee2021nnu} was trained out of the box (ResEncL architecture), but with 1\,mm spacing. Models Genesis and CLIP were pre-trained on all CTs and on all CT-Report pairs, respectively, and then fine-tuned on all CT-Mask pairs. MTL was trained jointly on all CT-Report and CT-Mask pairs. Classification was trained on all CTs, with tumor presence/absence labels (per organ) from reports. Segmentation (MedFormer) and nnU-Net were only trained on CT-Mask pairs. ULS, Merlin, and MedGemma used their public checkpoints.

\textbf{\method\ surpasses prior public AI in detecting esophagus, spleen and uterus tumors (Tab. \ref{tab:rt_super_results_int} and \ref{tab:rt_super_results_small_combined}).} \method\ substantially surpasses Merlin, MedGemma, and ULS, three leading public AI models. Merlin and MedGemma are VLMs. They generated reports and we evaluated these reports for tumor detection (following~\cite{bassi2025radgpt}). The VLM reports had low sensitivity, missing the tumor types we analyzed here. VLMs were surpassed by segmentation models, as in~\cite{bassi2025radgpt,chen2025vision}. Here, VLM performance was even lower than in~\cite{bassi2025radgpt}, because our tumor types are rarer and more difficult to detect on CT. ULS surpassed VLMs, but still had low sensitivity, likely because our tumor types are rare or absent in its training data. Also, ULS is trained on tumor crops only, which puts it at a disadvantage in tumor detection. \method\ surpassed all the public AI models.

\textbf{\method\ improves tumor detection and segmentation with few or no tumor masks.} Tab.~\ref{tab:rt_super_results_int} shows that \method\ surpassed 8 other training methods that also trained on our dataset, but were not designed to jointly leverage reports, longitudinal images, and multi-phase images. Although \method\ performed similarly to R-Super in tumor detection, it substantially surpassed R-Super in segmentation DSC when training without masks (Tab.~\ref{tab:rt_super_results_small_combined}). Tab.~\ref{tab:rt_super_results_small_combined} displays diverse ablation studies, showing how each component of \method\ improves performance.

\textbf{Conclusion.} Tumor mask creation is expensive and time-consuming, leaving public datasets and segmentation models unable to cover most tumor types. Following R-Super~\cite{bassi2025learning,bassi2025scaling}, \method\ shows that AI can learn tumor segmentation from routinely available hospital data (reports, longitudinal images, and multi-phase images) rather than from manual masks. This can scale tumor segmentation to more tumor types.

\begin{credits}
\subsubsection{\ackname} This work was supported by the Lustgarten Foundation for Pancreatic Cancer Research and the National Institutes of Health (NIH) under Award Number R01EB037669. Paper content is covered by patents pending. We would like to thank the Johns Hopkins Research IT team in \href{https://researchit.jhu.edu/}{IT@JH} for their support and infrastructure resources where some of these analyses were conducted; especially \href{https://researchit.jhu.edu/research-hpc/}{DISCOVERY HPC}. %We thank Jaimie Patterson for writing a \href{}{news article} about this project. 

\subsubsection{\discintname}
The authors declare no competing interests.
\end{credits}
% \clearpage
\bibliographystyle{splncs04}
\bibliography{refs,zzhou}

\end{document}